\documentclass[runningheads]{llncs}
\usepackage[T1]{fontenc}
\usepackage{graphicx}
\usepackage{pdfpages}
\usepackage{algorithm}
\usepackage{algorithmic}

\usepackage{booktabs}       % professional-quality tables
\usepackage{amsmath}
\usepackage{amssymb}
\usepackage[utf8]{inputenc} % allow utf-8 input
\usepackage[T1]{fontenc}    % use 8-bit T1 fonts
\usepackage{url}            % simple URL typesetting
\usepackage{nicefrac}       % compact symbols for 1/2, etc.
\usepackage{microtype}      % microtypography
\usepackage{xcolor}
\usepackage{enumitem}
\usepackage{subcaption}
\newtheorem{assumption}{Assumption}
\usepackage[misc]{ifsym}

\begin{document}

\pagestyle{empty}

\title{Less is More: Efficient Weight ``Farcasting'' with 1-Layer Neural Network}

% \titlerunning{Abbreviated paper title}
% If the paper title is too long for the running head, you can set
% an abbreviated paper title here

\author{Xiao Shou \Letter \inst{1}%\orcidID{0000-1111-2222-3333} 
\and
Debarun Bhattacharjya\inst{2}%\orcidID{1111-2222-3333-4444} 
\and
Yanna Ding\inst{3}%\orcidID{1111-2222-3333-4444} 
\and
Chen Zhao\inst{1}%\orcidID{1111-2222-3333-4444} 
\and
Rui Li\inst{4}%\orcidID{1111-2222-3333-4444} 
\and
Jianxi Gao\inst{3}%\orcidID{1111-2222-3333-4444} 
}
\authorrunning{Shou et al.}
% First names are abbreviated in the running head.
% If there are more than two authors, 'et al.' is used.
%
\institute{Baylor University, Waco, TX 76706, USA \\
\{Xiao\_Shou, Chen\_Zhao\}@baylor.edu
\and
IBM AI Research, Yorktown Heights, NY 10598 \\
\{debarunb@us.ibm.com\} \and
Rensselaer Polytechnic Institute, Troy, NY 12180 \\
\{dingy6,gaoj8\} @rpi.edu
\and Amazon \\
\{ruilit@amazon.com\}
}

\maketitle              % typeset the header of the contribution
\begin{abstract}
Addressing the computational challenges inherent in training
large-scale deep neural networks remains a critical endeavor
in contemporary machine learning research. While previous
efforts have focused on enhancing training efficiency through
techniques such as gradient descent with momentum, learning rate scheduling, and weight regularization, the demand for further innovation continues to burgeon as model sizes keep expanding. In this study, we introduce a novel framework
which diverges from conventional approaches by leveraging
long-term time series forecasting techniques. Our method capitalizes solely on initial and final weight values, offering a
streamlined alternative for complex model architectures. We
also introduce a novel regularizer that is tailored to enhance the
forecasting performance of our approach. Empirical evaluations conducted on synthetic weight sequences and real-world deep learning architectures, including the prominent large language model DistilBERT, demonstrate the superiority of our
method in terms of forecasting accuracy and computational
efficiency. Notably, our framework showcases improved performance while requiring minimal additional computational overhead, thus presenting a promising avenue for accelerating the training process across diverse tasks and architectures.

\end{abstract}
\section{Introduction}
\label{intro}
Large-scale deep learning systems such as large language models (LLMs) typically require significant investments of time and computation for training. For instance, GPT-3~\cite{brown2020language} demands several days of training even on high-performance machines, posing substantial challenges in terms of efficiency and resource allocation. 
%This immense time commitment poses substantial challenges in terms of efficiency and resource allocation. 
Over the years, numerous efforts have been dedicated to addressing the computational demands of training large-scale models \cite{menghani2023efficient}. 
Techniques such as architecture minimization and compression \cite{tan2019efficientnet} and the deployment of efficient physical hardware systems have yielded improvements in efficiency \cite{khailany2020accelerating}. 

From an algorithmic perspective, diverse training strategies aim to train the model in various ways, leading to fewer prediction errors, reduced data requirements, faster convergence, and other benefits. This enhanced quality can then be leveraged to create a smaller, more efficient model by trimming the number of parameters if needed. Two notable examples of optimization techniques include momentum-based methods like Adam and knowledge distillation-based methods like DistilBERT \cite{sanh2019distilbert,hinton2015distilling}. While these methods contribute to learning stability and marginally accelerate the training process, they do not directly address the fundamental issue of reducing overall training time.

Efforts to cut down training time typically  leverage short-term forecasting for predicting parameter weights; notable examples include \emph{Introspection}~\cite{sinha2017introspection} and the \emph{Weight Nowcaster Network (WNN)}~\cite{jang2023learning}. For instance, WNN incorporates two feed-forward neural networks for capturing both weight parameters and temporal differences to predict the weights of the target neural network over the next 5 steps, a technique termed ``nowcasting''. %However these 
Such approaches have several shortcomings: (i) they can only forecast in the short term, which limits their applicability to further reduce training time; (ii) they have only been applied to deep learning systems with fewer number of parameters -- in particular, WNN involves complex transformations %and computations 
that inhibit their application in LLMs as demonstrated later in our experiment on DistilBERT~\cite{sanh2019distilbert}; (iii) they can only forecast to 1 future time step which may not be optimal if abrupt changes occur in the training. 
%they can only forecast to 1 future time step which may not be optimal if abrupt changes are occurring in the training. 

We address limitation (i) by focusing on long-term weight prediction, which we term ``farcasting''. This concept is akin to long-term time series forecasting, where we predict many steps ahead. The motivation for adopting time series forecasting techniques in predicting neural network weights over time is driven by several key factors. Firstly, numerous state-of-the-art techniques have been developed for long-term time series forecasting% in time series forecasting problems
~\cite{zeng2023transformers}. The training process of neural networks, where weights and their updates can be seen as multivariate vectors or multivariate observations, shares similarities with time series data. However, a major distinction exists: neural network weight updates are governed by specific update rules, while time series data typically exhibit periodicity and trends. %To further improve on (ii) 
To address limitation (ii),
we explore the feasibility of using a simple model — a one-layer feed-forward neural network — for this forecasting task, which further reduces training time, marking a major leap forward in the efficiency of deep learning practices. By allowing our neural network to predict a sequence of future steps leveraging direct multi-step forecasting strategies~\cite{chevillon2007direct}, we naturally address limitation (iii) and enhance model stability by smoothing out abrupt parameter changes. 

Our \textbf{contributions} are summarized as follows: 
% (1) We introduce the concept of "farcasting" for parameter prediction in the training of machine learning (ML) and deep learning (DL) systems, emphasizing large-scale models.(2) We demonstrate that even a one-layer neural network can effectively solve linear systems, highlighting the potential for simplicity in forecasting. (3) We show that a much smaller feed-forward fully connected network can significantly reduce the number of parameters while achieving accuracies that are equivalent or superior to those from larger networks. (4) We present compelling applications in computer vision and LLMs, utilizing two notable architectures: convolutional neural networks (CNNs) and DistilBERT.

\begin{itemize} [noitemsep,nolistsep,leftmargin=*]
\item We introduce the concept of "farcasting" for parameter prediction in the training of machine learning (ML) and deep learning (DL) systems, emphasizing large-scale models.
%with a particular emphasis on large-scale models.
\item We demonstrate that even a one-layer neural network can effectively solve linear systems, highlighting the potential for simplicity in forecasting.
\item We show that a much smaller feed-forward fully connected network can significantly reduce the number of parameters while achieving accuracies that are equivalent or superior to those from larger networks.
\item We present compelling applications in computer vision and LLMs, utilizing two notable architectures: convolutional neural networks (CNNs) and DistilBERT.
\end{itemize}

%1) We introduce the concept of "farcasting" for parameter prediction in the training of machine learning (ML) and deep learning (DL) systems, with a particular emphasis on large-scale models. 2). We demonstrate that even a one-layer neural network can effectively solve linear systems, highlighting the potential for simplicity in forecasting. 3). We show that a much smaller feed-forward fully connected network can significantly reduce the number of parameters while achieving accuracies that are equivalent or superior to those from larger networks. 4). We present compelling applications in computer vision and LLMs, utilizing two notable architectures: convolutional neural networks (CNNs) and DistilBERT. %Our approach demonstrates the potential for significant advancements in both fields, paving the way for more efficient and scalable deep learning models.
% By addressing the critical issue of training duration, our work aims to set a new standard in the development and deployment of large-scale deep learning systems, offering a pathway to more efficient and sustainable AI solutions.

\section{Related Work}
\label{rel}

\paragraph{Parameter Prediction of Deep Learning.} 
% A closely related line of work to ours is to learning to boost deep learning training. A prior work Introspection network \cite{sinha2017introspection}, which utilizes the weight history of unseen neural networks and sample points from these weights to predict future values using another introspection network. This methodology accelerates the training process of the unseen neural network. 
% {\color{yd} Knyazev et al. \cite{knyazev2021parameter} expanded on Graph HyperNetworks~\cite{zhang2018graph}  by training a hypernetwork to predict a set of performant parameters for the source task. However, since these parameters are often suboptimal, they serve as initialization for further training.}
% In contrast, our approach diverges by forecasting the trajectory of weights rather than predicting a single point value for the neural network. This strategy completely circumvents the need for further training. Another relevant is 
% the Weight Nowcaster Network (WNN), a module that periodically nowcasts near future weights \cite{jang2023learning}. Orthogonal to these two works which only predicts a few steps further, we tackle these more aggressively by long term forecasting over many steps further, which will significant boost training of large scale deep learning systems.  
A closely related line of work to ours focuses on accelerating deep learning training. A notable prior work, the Introspection Network \cite{sinha2017introspection}, uses the weight history of unseen neural networks and sample points from these weights to predict future values using another introspection network, thereby speeding up the training process of the unseen neural network. Similarly, Knyazev et al. \cite{knyazev2021parameter} expanded on Graph HyperNetworks \cite{zhang2018graph} by training a hypernetwork to predict a set of performant parameters for the source task. However, since these parameters are often suboptimal, they serve as initialization for further training. In contrast, our approach diverges by forecasting the trajectory of weights rather than predicting a single point value for the neural network. This strategy largely circumvents the need for further training. Another relevant work is the Weight Nowcaster Network (WNN), which periodically nowcasts near-future weights \cite{jang2023learning}. Unlike these methods, which only predict a few steps ahead, our approach tackles long-term forecasting over many steps, significantly boosting the training of large-scale deep learning systems.

% \paragraph{Time Series forecasting.} Our work is also closely related to time series long-term forecasting. Many transformer based models have been proposed such as \cite{nie2022time,zhang2022crossformer,xue2023make}. While those models provide state-of-the-art predictive performance, the computation of transformer may be prohibitive in our setting. Others have used foundation model approaches \cite{ye2024survey}. A feed forward neural network alternative are more feasible in our setting. We see previous work on this subject are DLinear and its variant from \cite{zeng2023transformers} and with some transformation from \cite{li2023revisiting} to capture periodictity and trends. However the periodsdictity may be irrelavant in our setting as weight sequences do not hold this property as seasaonality. 

\paragraph{Time Series Forecasting.} Our work is closely related to long-term time series forecasting. Various transformer-based models, such as \cite{xue2023make,nie2022time}, offer state-of-the-art predictive performance. However, the computational demands of transformers may be prohibitive in our setting. Alternatively, foundation model approaches have been explored, as noted by Ye et al. \cite{ye2024survey}. In our context, feed-forward neural networks present a more feasible option. Previous relevant work includes DLinear and its variants by Zeng et al. \cite{zeng2023transformers}, as well as methods incorporating transformations to capture periodicity and trends, as discussed in a more recent work \cite{li2023revisiting}. However, periodicity may be irrelevant in our setting since weight sequences do not exhibit periodic properties.

\paragraph{ODE Forecasting.} Numerous methods have been proposed for time series forecasting, utilizing convolutional networks, neural Ordinary Differential Equations (ODEs) \cite{chen2018neural}, and transformers. Temporal convolutional neural networks \cite{luo2024moderntcn} hierarchically aggregate adjacent timestamps to capture time series patterns. However, these approaches often rely on previously predicted data for future value rollouts, leading to inaccuracies in long-term forecasting due to error propagation. Neural ODEs fit observed data to a system of differential equations, while Graph ODEs \cite{zang2020neural,huang2020learning,huang2021coupled,luo2023hope,ding2024architecture} extend this concept to model coupled dynamical systems. Nevertheless, training differential equation models involves computationally intensive numerical integration, especially for large networks. In this study, we investigate discrete time forecasting approaches to address the modeling of parameter trajectories.

\paragraph{Large Scale Optimization.} Modern large-scale optimization and deep learning primarily utilize first-order methods, such as mini-batch stochastic gradient descent and its momentum-based variants like Adam \cite{kingma2014adam} and AdamW \cite{loshchilov2017decoupled}, to achieve faster convergence. Additional techniques include learning rate scheduling \cite{he2016deep,loshchilov2016sgdr}, applying regularizers to trainable parameters in networks \cite{rodriguez2016regularizing,jia2018highly}, and various initialization strategies \cite{glorot2010understanding,he2015delving}. Some researchers have also explored optimization algorithms by framing it as learning to learn an optimizer \cite{andrychowicz2016learning}. Our approach is orthogonal to these efforts; we instead frame the problem as a forecasting task.

% \paragraph{Large Scale Optimization.} Modern large scale optimization and deep learning rely on first order methods such as mini batch stochastic gradient descent or its moment based variant such as Adam \cite{kingma2014adam} and AdamW \cite{loshchilov2017decoupled} for faster convergence. Other techniques are scheduling of learning rate \cite{he2016deep,loshchilov2016sgdr}, applying regularizers on
% trainable parameters in networks \cite{rodriguez2016regularizing, jia2018highly}, and various initialization techniques have been developed \cite{glorot2010understanding,he2015delving}. Others investigate optimization algorithms and cast it as learning to learn an optimizer \cite{andrychowicz2016learning}. Ours are orthogonal to their work; we cast as a forecasting task. 

% \paragraph{ODE forecasting} % should we merge ODE forecasting to time-series long term forecasting? 

% \paragraph{meta learning?}

\section{Linear Regression: An Example}
In large-scale machine and deep learning endeavors, the primary goal is to minimize a designated loss function $L$ concerning certain parameters $\phi$ within a defined parameter space. A prevalent optimization technique in deep learning is mini-batch stochastic gradient descent (SGD), where a sequence of weights and biases in $\phi$ and represented as $(\mathbf{w}_1, \mathbf{w}_2,...,\mathbf{w}_n)$ (where $\mathbf{w}_i$ includes both weights and biases)
 %, as $(\mathbf{w}_1, \mathbf{w}_2,...,\mathbf{w}_n)$,
undergoes $n$ iterations of weight updates: $\mathbf{w}_{i+1} \leftarrow \mathbf{w}_i - \alpha \nabla L(\mathbf{w}_i)$. 

For instance, consider a typical problem in machine and deep learning, such as linear regression, a common statistical and machine learning task. The objective is to minimize the sum of squared errors. Given a data matrix $\mathbf{D} \in \mathbb{R}^{n \times d}$ comprising $n$ samples, each with $d$ features, and a response vector $\mathbf{e} \in \mathbb{R}^n$ containing $n$ responses, the aim is to find a weight vector $\mathbf{w} \in \mathbb{R}^d$ that minimizes the squared loss between predicted responses and the ground truth:

\begin{equation}
\label{eqn:lsm}
\min_\mathbf{w} \frac{1}{2}|| \mathbf{D}\mathbf{w}-\mathbf{e} ||_2^2
\end{equation}

This represents an unconstrained quadratic problem with a unique optimal solution $w^* = (\mathbf{D}^\intercal \mathbf{D})^{-1} \mathbf{D}^\intercal \mathbf{e}$. Gradient descent, a common optimization method, iteratively generates a sequence of $\mathbf{w}_i$'s using updates of the form $\mathbf{w}_{i+1} \leftarrow \mathbf{w}_{i} -\alpha \nabla L(\mathbf{w}_{i}) $, where $\nabla L(\mathbf{w}_i) =(\mathbf{D}^\intercal \mathbf{D}) \mathbf{w}_i - \mathbf{D}^\intercal \mathbf{e}$ for the quadratic optimization problem.

In contrast, a feed forward neural network $g_\theta (\cdot)$ offers a more expressive modeling approach, aiming to minimize the sum of squared errors between the predicted and ground truth parametrized by weights and bias in a neural network:
\begin{equation}
\label{eqn:lsnn}
\min_\theta \frac{1}{n} || g_\theta(\mathbf{D})-\mathbf{e} ||_2^2
\end{equation}
Here, $ g_\theta(\mathbf{D}) = \sigma_k(... \sigma_2( \sigma_1(\mathbf{D} \cdot \mathbf{W}_1 + \mathbf{b}_1) \cdot \mathbf{W}_2 + \mathbf{b}_2 ...) $ represents a $k$-layer neural network with weights $\mathbf{W}_i$ and biases $\mathbf{b}_i$, and activation functions $\sigma_i$ such as RELU. For the above minimization problem, the final activation is identity. Notably, this problem is generally non-convex, lacking a guaranteed unique optimal solution. Mini-batch stochastic gradient descent optimizer or one of its variants is typically employed for training the network. %We will use the two cases of linear regression in the following sections to illustrate our approach and perform synthetic experiments.
We will consider the two afore-mentioned cases of regression in the following sections to illustrate our approach through synthetic data experiments.

\section{Method}
\label{med}
\begin{figure}
    \centering
    \includegraphics[width=0.9\linewidth]{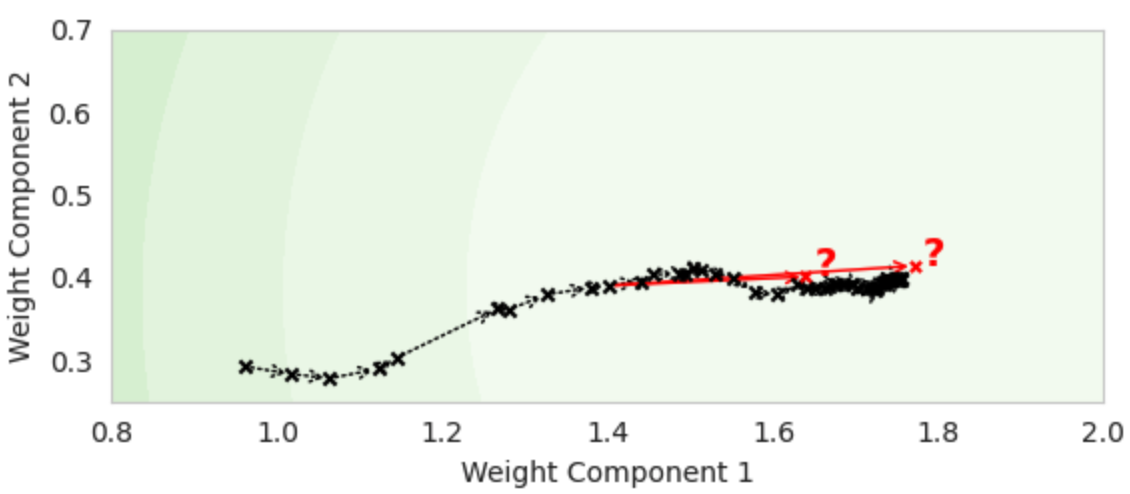}
\caption{Illustration of the weight forecasting procedure. Weight sequences are generated for a two-dimensional linear regression problem from Equation~\ref{eqn:lsm}. 
Black points are weights from SGD, red points are predictions from forecasting. The red arrow \textcolor{red}{$\rightarrow$} shows farcasting from a partial weight sequence.
%x: Weights by SGD. \textcolor{red}{x}: Farcasted Weights. ...: SGD Update. 
%\textcolor{red}{$\rightarrow$}:Farcasting. 
}
    \label{fig:lr}
\end{figure}

Analogous to long-term time series forecasting (LTSF), we are given a sequence of weights from (stochastic) gradient descent or one of its variants of $n+1$ steps, denoted as $ \mathbf{X} = (\mathbf{w}_0, \mathbf{w}_1,\mathbf{w}_2,...,\mathbf{w}_n)$ and our goal is to predict future weights $m$ steps ahead, denoted as $ \mathbf{Y} = (\mathbf{w}_{n+1},\mathbf{w}_{n+2},...,\mathbf{w}_{n+m})$, where $m \gg 10$. For simplification, our prediction can be expressed as $ \mathbf{X} \in \mathbb{R}^{d \times (n+1)} \rightarrow \mathbf{Y} \in \mathbb{R}^{d \times m}$. 
Figure \ref{fig:lr} illustrates our problem using 2 dimensional linear regression.

We pursue the direction of direct multi-step forecasting \cite{zeng2023transformers} instead of iterated multi-step (IMS) \cite{taieb2012recursive} due to more error accumulation from each autoregressive step in IMS.
We emphasize that here %our pair 
$(\mathbf{X}, \mathbf{Y})$ specifically denotes the input output pairs which are parameter (sub)sequences from training or optimization processes. The ultimate objective is to significantly reduce optimization steps, crucial for large-scale machine and deep learning systems. While there are many state-of-the-art transformer based models for LTSF such as \cite{nie2022time}, they are not suitable here as the computational overhead for training such large models may well be greater than performing stochastic gradient update for the current system. We thus restrict our attention to feed forward neural networks. In particular, we are interested in finding a neural network $h \in \mathcal{H}$ in the space of all neural networks that maps from a sequence of weights $\mathbf{X} \in \mathbb{R}^{d \times (n+1)}$ from (stochastic) gradient descent or its variants to a $\mathbf{Y} \in \mathbb{R}^{d \times m}$. For a simple 1 layer neural network $h_{\theta}: \mathbf{X} \rightarrow \mathbf{Y}$ where parameter set $\theta$ contains weights and bias $\{\mathbf{A}, \mathbf{b}\}$, we aim to find a solution for the system of linear equations: $ \mathbf{X} \cdot \mathbf{A} + \mathbf{b} =  \mathbf{Y}$. In essence, our method learns a mapping from past weight sequences to future weights using a simple feedforward layer, enabling multi-step forecasting and reducing the need for backpropagation at every step. For a simplified system, we arrive at the following statement. 

\begin{proposition}
\label{theo}
If the (stochastic) gradient descent (or its variant) update (i.e., $\mathbf{w}_{i+1} \leftarrow \mathbf{w}_i - \alpha \nabla L(\mathbf{w}_i)$) can be written as $\mathbf{w}_{i+1} = c_i \cdot \mathbf{w}_i + d_i$ for some $c_i, d_i \in \mathbb{R}$, then there exists $\mathbf{A}$ and $\mathbf{b}$ that solve  $\mathbf{X} \cdot \mathbf{A} + \mathbf{b} =  \mathbf{Y}.$
\end{proposition}

Proof sketch. For this linear system of equations, we have $[ \mathbf{w}_1,\mathbf{w}_2,...,\mathbf{w}_n] \cdot \mathbf{A}_i + \mathbf{b}_i = \mathbf{w}_{n+i} $
for $i \in \{1,2,...,m\}$, or more compactly $\sum_{j=1}^n \mathbf{w}_j \cdot \mathbf{A}_{ji} + \mathbf{b}_i = \mathbf{w}_{n+i}$. We solve for the case where $i=1$, and then $i=2$ and so on. $\sum_{j=1}^n \mathbf{w}_j \cdot \mathbf{A}_{j1} + \mathbf{b}_1 = \mathbf{w}_{n+1}$. Note we also have $c_n \cdot \mathbf{w}_n + d_n = \mathbf{w}_{n+1}$. We can choose $\mathbf{A}_1$ to be some scalar multiplied by a one-hot vector with 1 on the last term, i.e. $\mathbf{A}_1 = c_n \cdot [0,0,0,...,1] ^\intercal$ and $\mathbf{b}_1 = d_n$. Similarly we can solve for the $i=2$ case by noting $\mathbf{w}_{n+2} = c_n c_{n+1} \cdot \mathbf{w}_{n} + c_{n+1} d_n + d_{n+1}$ using our recurrence relation. Hence $\mathbf{A}_2 =  c_n c_{n+1}  \cdot [0,0,0,...,1] ^\intercal $ and $\mathbf{b}_2 = c_{n+1} d_n + d_{n+1}$. We can perform this procedure until $m$ and then acquire $\mathbf{A}^*$ to be a matrix such that the only row that contains nonzero entries - the ${n}^{th}$ row - contains scalar multiples and the $i^{th}$ entry is  $c_{n} \cdot c_{n+1} \cdot ... c_{n+i-1}$ for $i \in \{1,2,3,...n\} $.
%\qed 
\hfill
$\square$

While the assumption that the update is linear is a simplification in modern large scale deep learning systems, it serves as a linear approximation to many complex systems. More significantly, a simple solution $\mathbf{A}^*$ places the nonzero entries on the $n^{th}$ row, which inspires our approach to design more efficient farcasting techniques. In practice, when we have a dataset with $l$ pairs of $\mathbf{X}$ and $\mathbf{Y}$, we seek a computation-friendly numerical optimization approach using neural networks to minimize the prediction error: 
\begin{equation}
\label{eqn:linear}
\min_{\{\mathbf{A},\mathbf{b}\}} \frac{1}{l}\sum_{i=1}^{l} || \mathbf{X}^{i} \cdot \mathbf{A} + \mathbf{b} - \mathbf{Y}^i ||_{l_1}
\end{equation}
The entry-wise $l1$-norm is applied here due to the parameters from (stochastic) gradient descent or its are common larger at early steps are usually smaller at late ones and the prediction error should not be dominated by early ones. Another more involved approach is to employ some sort of preprocessing $P$ or transformation $\psi$ \cite{li2023revisiting} on $\mathbf{X}$. We use preprocessing to indicate such process is non-trainable, and transformation  for trainable approaches. They can be expressed as $\min_{\{\psi, \mathbf{A},\mathbf{b}\}} \frac{1}{l}\sum_{i=1}^{l}|| \psi(P(\mathbf{X}^i)) \cdot \mathbf{A} + \mathbf{b} - \mathbf{Y}^i ||_{l_1}$. While this approach may working reasonably well in time series forecasting \cite{li2023revisiting}, the sequential nature of parameter updates via (stochastic) gradient descent does not imply any composition of periodicity and trends commonly observed in time series data. Furthermore, such processing and transformation increases computational cost and leads to challenges in training the network as illustrated later by our experiments on DistilBERT. % in the section on experiments.%Section \ref{exp}.

\paragraph{Sampling Weights for Efficiency.}
A potential approach to further reduce computation is to use a subset of parameter vectors by constraining the previous linear objective function:
\begin{equation}
\label{eqn:LFS}
\begin{aligned}
\min_{\{\mathbf{A},\mathbf{b}\}} \quad & \frac{1}{l} \sum_{i=1}^{l}|| \mathbf{X}^{i}_T \cdot \mathbf{A} + \mathbf{b} - \mathbf{Y}^i ||_{l_1}  \\
\textrm{s.t.} \quad & T \subset \{0,1,2,3,...,n\} \\
\end{aligned}
\end{equation}
where $\mathbf{X}_T^i$ is the (sampled) subset of columns from $\mathbf{X}^i$. If we restrict the cardinality to $|T| = k$, we can achieve this by drawing $k$ columns. Much research has been conducted in this area, known as the column subset selection problem \cite{wang2018provably}, which often involves column importance sampling algorithms such as LinearSVD and score sampling. However, in our setting, performing such decomposition procedures for large $d$ (e.g., $10^8$) is costly. A simple approach is to draw $k$ columns without replacement from a uniform distribution over (0, n) as a preprocessing step.

%We design a procedure similar to Introspection network \cite{sinha2017introspection} that works empirically well with sample of 4. We use algorithm \ref{alg:lfs} for training. 

% \begin{algorithm}[tb]
%    \footnotesize
%    \caption{ Long-term Forecasting by Sampling }
%    \label{alg:lfs}
% \begin{algorithmic}
%    \STATE {\bfseries Input:} Sequences of Weights $W_1$,  $W_2$. 
%    \STATE Sample t $\sim$ Uniform(n/2,n+1)
%    \STATE Select $W_T^1$ = [$w_0$, $w_{7t/10}$,$w_{14t/10}$,$w_t$]
%    \STATE Train LFS via Equation \ref{eqn:LFS} \\
% \end{algorithmic}
% \end{algorithm} 

\paragraph{Importance of Weights for Learning and Prediction.}
Upon further reflection on the update rules commonly used in 
relevant
%machine learning and deep learning
optimizers, we observe that the most critical weight for predicting the weight at step $n + 1$ is the weight $\mathbf{w}_n$ from step $n$, followed by the weight $\mathbf{w}_{n-1}$ from step $n-1$, and so on. This reliance on the most recent weight is a hallmark of first-order gradient-based optimizers commonly used in deep learning, such as SGD and Adam \cite{kingma2014adam}. Additionally, we note that the solution $\mathbf{A}^*$
in Theorem~\ref{theo} assigns nonzero values only to the last term, further validating our proposed approach.  Our method is straightforward and `deterministic' — it selects the most recent weight $\mathbf{w}_n$, which is most relevant for future updates in our forecasting problem. Furthermore, we include the initial weight, as it determines the starting point and has been crucial in the study of weight initialization in deep learning~\cite{glorot2010understanding,he2015delving}. This importance is also reflected in our empirical investigations, demonstrating the significance of initial and final weights. % in training and prediction. 
Our approach offers the side benefits of reducing memory usage, runtime computation, and the number of parameters from the original $\mathcal{O}(nd)$ in Equation \ref{eqn:linear} to  $\mathcal{O}(kd)$ in Equation \ref{eqn:LFS}, and finally to $\mathcal{O}(d)$ in Equation \ref{eqn:LFD} below. 
%We provide additional information regarding computational reductions through our model in the Appendix.
% efficiency -> calculate number of flops for each cases. 
\begin{equation}
\label{eqn:LFD}
\min_{\{\mathbf{A},\mathbf{b}\}} L_{pred}( \{\mathbf{A},\mathbf{b}\} ) = \frac{1}{l}\sum_{i=1}^{l}|| \mathbf{X}_{[0,n]}^i \cdot \mathbf{A} + \mathbf{b} - \mathbf{Y}^i ||_{l_1}  
\end{equation}

\paragraph{Background Knowledge from First Order Optimality Condition.} The training process for most DL systems is largely non-convex, where a typical model converges to a (local) optimal. This naturally leads to the following assumption.

\begin{assumption}
\label{assum:stop}
%The weights are updated via (stochastic) gradient descent (or its variant), i.e., $\mathbf{w}_{i+1} \leftarrow \mathbf{w}_i - \alpha_i \nabla L(\mathbf{w}_i)$). 
According to  the first order optimality condition for unconstrained optimization problems, the sequence of weights updated via (stochastic) gradient descent (or its variant) follows $\lim_{i \rightarrow \infty} ||\nabla L(\mathbf{w}_i)||_{l_p} = 0$ with respect to some norm $1 \le l_p \le \infty$.
\end{assumption}

From Assumption \ref{assum:stop}, we can deduce $||\nabla L(\mathbf{w}_n)||_{l_1} \le ||\nabla L(\mathbf{w}_0)||_{l_1}$ for some large $n$.  Meanwhile we have $\mathbf{w}_{i+1} - \mathbf{w}_i = - \alpha_i \nabla L(\mathbf{w}_i)$. Commonly in ML/DL, the learning rate remains constant or decays according to some annealing procedure such that $0 \le \alpha_j \le \alpha_0$ for $j \in \{1,2,3,...,n\}$. We arrive at $|| \mathbf{w}_1 - \mathbf{w}_0 ||_{l_1} \ge  || \mathbf{w}_{n+1} - \mathbf{w}_n ||_{l_1} $ for some large $n$. Hence we impose a soft penalty $L_{grad}$ to incorporate such knowledge for any deviation from this prediction: 
\begin{equation}
L_{grad} =   \sum_{j}^{J}  \max(|| \hat{\mathbf{w}}_{j+1} - \hat{\mathbf{w}}_j ||_{l_1} - || \mathbf{w}_1 - \mathbf{w}_0 ||_{l_1}, 0 ) 
\end{equation}
for some large $j$ and $\hat{\mathbf{w}}_j
$'s are predicted weights from Equation \ref{eqn:LFD}. In practice, we can choose $j$ to be $n+m-1$. Our proposed linear farcaster which includes the first and last weight steps (referred to as \textbf{LFD-2}), aims to minimize the combined loss $L_{pred} + \beta \cdot L_{grad}$, where $\beta$ balances the two losses; $\beta$ is typically set to a small value, especially when the dimension $d$ is large.

\paragraph{Remark on Efficiency.} We demonstrate the efficiency of our approach via the computing of number of FLOPs. Consider performing $m$ updates of $\mathbf{w}_{i+1} \leftarrow \mathbf{w}_i - \alpha_i \nabla L(\mathbf{w}_i)$, where $\mathbf{w}_i \in \mathbb{R}^d$ and $d \gg 1$. In a simple case where the gradient is linear, $\nabla L(\mathbf{w}_i) = \mathbf{C} \mathbf{w}_i - \mathbf{h}$, for matrix $\mathbf{C}$ and vector $\mathbf{h}$, each update operation requires $2d^2+2d$ FLOPs. Consequently, $m$ updates lead to $2md^2+2dm$ FLOPs. For our trained model LFS-2 at inference phase, finding $m$ updates demands only $4dm$ FLOPs, while retrieving the $m^{th}$ update alone requires only $4d$ FLOPs which results in significant gains in efficiency. Large-scale deep learning systems frequently involve more complex updates and thus any iterative update scheme will require more FLOPs.

\section{Experiments}
\label{exp}
We implement our approach using PyTorch \footnote{Our code is publicly available at \\ https://github.com/xshou1990/Less\_is\_more\_weight\_farcasting} and evaluate its performance on both synthetic and real-world weight sequences. The synthetic sequences are derived from our linear regression example, while the real-world sequences are obtained from training two representative deep learning models: a convolutional neural network and DistilBERT. Additional implementation details and training code are provided in Section 1 of the Appendix.

\begin{table} [hbtp]
\setlength{\tabcolsep}{2.5pt}
  \caption{Summary of baseline models. }
  \centering
  \resizebox{0.9\linewidth}{!}{
  \begin{tabular}{lll}
\toprule
   Model  &  Functional Form & Output  \\
    \midrule
   Introspection &    Neural Network $f: \mathbf{X}_T \in \mathbb{R}^{d \times 4} \rightarrow \mathbf{y} \in \mathbb{R}^{d}$ & Point \\
    WNN &  Neural Network $ f: \psi ([\mathbf{X}, d\mathbf{X}]) \in \mathbb{R}^{d \times l} \rightarrow y \in \mathbb{R}^{d} $ & Point
  \\
LFN     &  Neural Network $f: \mathbf{X} \in \mathbb{R}^{d \times n} \rightarrow \mathbf{Y} \in \mathbb{R}^{d \times m} $ & Sequence
\\
   DLinear & Neural Network $f : \psi(\mathbf{X}) \in \mathbb{R}^{d \times l} \rightarrow \mathbf{Y} \in \mathbb{R}^{d \times m} $  & Sequence
    \\
   LFS & Neural Network $f: \mathbf{X}_T \in \mathbb{R} ^{d \times k} \rightarrow \mathbf{Y} \in \mathbb{R}^{d \times m}$  & Sequence  \\
   LFL & Neural Network $f: \mathbf{X}_n \in \mathbb{R} ^{d \times 1 } \rightarrow \mathbf{Y} \in \mathbb{R}^{d \times m} $   & Sequence\\
  LFD-2 & Neural Network $f: \mathbf{X}_{[0,n]} \in \mathbb{R} ^{d \times 2} \rightarrow \mathbf{Y} \in \mathbb{R}^{d \times m} $   & Sequence \\
    \bottomrule
  \end{tabular}
}
   \label{tab:method}
\end{table}  

% \begin{figure*} [htbp]
%     \centering
%     \begin{subfigure}[b]{0.25\textwidth}
%         \centering
%         \includegraphics[width=\textwidth]{figs/3gd_new.png}
%         \caption[Network2]%
%         {{\small Syn-1 GD}}    
%         % \label{fig:mean and std of net14}
%     \end{subfigure}
%     \hfill
%     \begin{subfigure}[b]{0.24\textwidth}  
%         \centering 
%         \includegraphics[width=\textwidth]{figs/3sgd_new.png}
%         \caption[]%
%         {{\small Syn-1 SGD}}    
%         % \label{fig:mean and std of net24}
%     \end{subfigure}
%    % \vskip\baselineskip
%     \begin{subfigure}[b]{0.238\textwidth}   
%         \centering 
%         \includegraphics[width=\textwidth]{figs/31sgd_new.png}
%         \caption[]%
%         {{\small Syn-2 SGD}}    
%         % \label{fig:mean and std of net34}
%     \end{subfigure}
%     \hfill
%     \begin{subfigure}[b]{0.238\textwidth}   
%         \centering 
%         \includegraphics[width=\textwidth]{figs/31adam_new.png}
%         \caption[]%
%         {{\small Syn-2 Adam}}    
%         % \label{fig:mean and std of net44}
%     \end{subfigure}
% \caption[ The average and standard deviation of critical parameters ]
% {\small Examples of various generated weight sequences for the synthetic data experiments. The feature vectors are of 3 dimensions for the Syn-1 and 31 for Syn-2. We plot the values for each component over 200 updates in subplot (a)-(d) respectively.} 
% \label{fig:tru}
% \end{figure*}

\subsection{Synthetic Data Experiments} 
\paragraph{Synthetic Weight Sequence Generation.}
We prepare two types of synthetic data to validate our approach:
\begin{itemize}
\item \textbf{Syn-1}: We generate weight sequences from the least square regression problem in Equation \ref{eqn:lsm}. Specifically, we randomly sample optimal weight $\mathbf{w}^* \in \mathbb{R}^3$ from a normal distribution $\mathcal{N}(0, \mathbf{I})$. We similarly sample 100 feature vectors from $\mathcal{N}(0, \mathbf{I})$ to form a feature matrix $\mathbf{D} \in \mathbb{R}^{100 \times 3}$. Correspondingly the generated response vector $\mathbf{e}$ can be computed according to $\mathbf{D} \mathbf{w}^* = \mathbf{e}$. This procedure is similar to previous work on in-context learning \cite{garg2022can}. Once we have feature-response pair ($\mathbf{D}$, $\mathbf{e}$), we can formulate a linear regression problem to minimize the least square error.  Weight sequences $\mathbf{w}_i$ are generated according to gradient descent and mini-batch stochastic gradient descent respectively. %This is our synthetic experiment 1 or Syn-1 for short.
\item \textbf{Syn-2}: We similarly generate feature-response pair ($\mathbf{D} \in \mathbb{R}^{100 \times 1}$, $\mathbf{e \in \mathbb{R}^{100} }$) where this time the regression is performed via a 2-layer fully connected neural network with a total of 31 parameters to minimize, according to Equation \ref{eqn:lsnn}. We then obtain parameter sequences $\mathbf{w}_i \in \mathbb{R}^{31}$ from training the neural network. Optimizers SGD and Adam are applied to train the neural network respectively. %This is considered as synthetic experiment 2 or Syn-2. 
\end{itemize}

%We generate weight sequences from the least square regression problem according to Equation \ref{eqn:lsm}. Specifically, we randomly sample optimal weight $\mathbf{w}^* \in \mathbb{R}^3$ from a normal distribution $\mathcal{N}(0, \mathbf{I})$. We similarly sample 100 feature vectors from $\mathcal{N}(0, \mathbf{I})$ to form a feature matrix $\mathbf{D} \in \mathbb{R}^{100 \times 3}$. Correspondingly the generated response vector $\mathbf{e}$ can be computed according to $\mathbf{D} \mathbf{w}^* = \mathbf{e}$. This procedure is similar to previous work in in-context learning \cite{garg2022can}. Once we have feature-response pair ($\mathbf{D}$, $\mathbf{e}$), we can formulate a linear regression problem to minimize the least square error.  Weight sequences $\mathbf{w}_i$ were generated according to gradient descent and mini-batch stochastic gradient descent, respectively. This is our synthetic experiment 1 or Syn-1 for short. 

%We similarly generate feature-response pair ($\mathbf{D} \in \mathbb{R}^{100 \times 1}$, $\mathbf{e \in \mathbb{R}^{100} }$). This time the linear regression is performed via a 2-layer fully connected neural network with a total of 31 parameters to minimize according to Equation \ref{eqn:lsnn}. We then obtain the parameter sequences $\mathbf{w}_i \in \mathbb{R}^{31}$ from training the neural network. Optimizers SGD and Adam are applied to train the neural network respectively. This is considered as synthetic experiment 2 or Syn-2. 

We generate 200 minimization problems for each synthetic experiment and thus 200 weight sequences are collected. Each one contains 201 time steps including the randomly initialized weight. %Four representative weight sequences are shown in Figure \ref{fig:tru}.
We carefully choose the learning rate so that it is not so large that it diverges, or so small that it hardly converges. For Syn-1, in the GD update case, we use the reciprocal of the largest eigenvalue of the Hessian matrix multiplied by $0.01$; in the SGD update case, we use a batch size of 8 and learning rate of 0.001. For Syn-2, in the SGD case, we set the learning rate of 0.002; and for Adam, 0.005; batch size of 64 is used for both experiments. 
We split the sequences into 100/50/50 as train/dev/test. For each sequence, we use the first 21 time steps of weights as training sequence $\mathbf{X}^{i}$ and the remaining 180 time steps for prediction  $\mathbf{Y}^{i}$. A total of 5 trials for each experiment are recorded. %Details around choosing an appropriate learning rates are discussed in Section 1 of the Appendix.

\paragraph{Baselines.}
We compare our model LFD-2 with 6 baseline models.  A summarized overview of each model is presented in Table \ref{tab:method}. Introspection \footnote{https://github.com/muneebshahid/introspection}~\cite{sinha2017introspection} and WNN \footnote{https://github.com/jjh6297/WNN}~\cite{jang2023learning} are weight forecasting modules that train on weight sequences and predict a near future weight, thus providing a single time step prediction. DLinear \footnote{https://github.com/plumprc/RTSF/tree/main} \cite{zeng2023transformers} is a powerful time series long-term forecasting model which was originally proposed to examine the necessity of applying complex transformer models in time series forecasting tasks; a few trainable transformations including MLP and RevIN \cite{kim2021reversible} are applied to better extract temporal and channel-wise features~\cite{li2023revisiting}. 

We also utilize simple baseline models known as LFN and LFL. LFN is a one-layer fully connected neural network designed to minimize Equation \ref{eqn:linear} by considering all steps in a sequences, while LFL uses solely the very last weight. Additionally, LFS draws inspiration from Introspection by selecting a subset of columns and minimizing the prediction error for each entry based on Equation \ref{eqn:LFS}. It is worth noting that, unlike Introspection and WNN, these models produce a sequence of weights for a single inference instance.

\paragraph{Evaluation Metrics.} While it is not straightforward to compare the trajectory against some ground truth, we provide checks at various checkpoints for weight step $i \in \{40, 80, 160, 200\}$ for quantitative evaluation similar to that of time series forecasting using mean squared error (MSE) \cite{nie2022time,zeng2023transformers}. For qualitative visual evaluation, we provide predicted trajectories for models other than WNN and Introspection, which is not uncommon in evaluating models for time series forecasting tasks. 

\begin{table*} [h!]
\tiny
\setlength{\tabcolsep}{2pt}
  \caption{MSE for various models at checkpoints, scaled by $10^4$ on Syn-1. Best results are in bold; second best ones are in italics. Standard deviation values are included in parentheses. }
  \centering
  \resizebox{1\linewidth}{!}{
  \begin{tabular}{lllllllll}
\toprule
 &  \multicolumn{4}{c}{GD} & \multicolumn{4}{c}{SGD} 
\\\cmidrule(r){2-5}\cmidrule(l){6-9} 
   Model/Time   & 40  & 80  & 160 & 200  & 40  & 80  & 160 & 200 \\
    \midrule
   Introspection & .222(.023) &	\textit{4.63(.41)} &	45.1(3.8) &	405(724) & 12.8(1.6) & 	\textit{61.4(13.7)}	& 219(48) &	307(45)
\\
   WNN & 1.75(1.76)	& 15.9(8.2)	& 	213(62)	& 278(60) &
14.1(.6) & 71.8(14.8)	& 270(29)		& 412(60)
  \\
LFN     &.619(.561)		& 5.11(1.18)	& \textbf{43.8(4.5)} & 92.4(12.0) &
 \textit{12.7(2.0)}	& 61.7(9.8)	& \textit{212(25)} &	339(44)
\\
   DLinear & .715(.119) 	&5.72(.42) & 48.1(3.9)	& \textit{81.8(6.8)}
& 14.2(2.3) &	68.6(13.6) &	221(43) &	\textit{302(53)}
    \\
   LFS & \textit{.373(.062)}	& 5.34(.65) & 60.3(18.7) &	205(112) & 103(118)	& 513(536) &	1557(1517)	& 2025(1917)
   \\
   LFL &193(23)&	1337(257)  &3787(353)  &	4594(181) & 177(10) &	1241(186) &	3615(203)& 4667(345) 
   \\
  LFD-2  & \textbf{.207(.021)} &	\textbf{4.39 (.39)}&	\textit{44.9(5.0)} & \textbf{81.1(9.6)} & \textbf{11.0(1.4)} &	\textbf{53.0(8.6)}	 & \textbf{195(22)}	&\textbf{278(18)}
			\\
    \bottomrule
  \end{tabular}
}
\label{tab:syn1}
\end{table*}

\begin{table*} [h!]
\tiny
\setlength{\tabcolsep}{2pt}
  \caption{%MSE for Various Models at Checkpoints, Scaled by $10^4$ on Syn-2. 
  MSE for various models at checkpoints, scaled by $10^4$ on Syn-2. Best results are in bold; second best ones are in italics. Standard deviation values are included in parentheses.
  }
  \centering
  \resizebox{1\linewidth}{!}{
  \begin{tabular}{lllllllll}
\toprule
  \multicolumn{5}{c}{SGD} & \multicolumn{4}{c}{Adam} 
\\\cmidrule(r){2-5}\cmidrule(l){6-9} 
   Model/Time   & 40  & 80  & 160 & 200  & 40  & 80  & 160 & 200 \\
    \midrule
   Introspection & 13.0(.8) &	55.6(40.3) &	158(51) &234(50) &12.5(4.9)& 53.1(19.9)	&85.3(21.4)	& 104(39)
\\
   WNN & \textbf{1.30(.20)}	& \textbf{21.9(2.0)}	& \textbf{109(8)} &	176(66)& 4.52(3.53)	& \textbf{9.27(2.26)}& \textit{44.4(27.9)} & 74.0(47.8)	
  \\
LFN     &2.20(.50) & \textit{25.6(3.50)} & 123(13)&173(20) & 7.88(3.42)  &	42.6(19.8)& 103(25)	& 129(30) 
\\
   DLinear & 4.99(.21) & 31.6(2.3)	& 126(9) &	\textbf{168(13)} & 2.60(.25) & \textit{12.7(.9)} & \textbf{37.2(3.4)} & \textbf{45.6(4.9)}
    \\
   LFS &5.41(4.59)&43.0(22.0)	&166(37)	&207(39) &11.1 (5.3)	&54.7 (13.6)	&108(22)&122(24)
   \\
   LFL & 8.97(.49)	&61.7(5.5) &195 (22)	&243(27)& 7.68(1.86)	&43.2(9.7) &101(16)&117(17)
   \\
  LFD-2 & \textit{1.83(.40)}	&26.2(3.6)&	\textit{123(12)}	&\textbf{168(21)} &
  \textbf{1.13 (.65)} &	16.3 (11.0) &	49.6 (20.4)	& \textit{60.2(11.1)}	
			\\
    \bottomrule
  \end{tabular}
}
\label{tab:syn2}
\end{table*}  

\paragraph{Results.} The results of the synthetic experiments Syn-1 and Syn-2 are presented in Tables \ref{tab:syn1} and \ref{tab:syn2}, respectively. For Syn-1, our proposed model LFD-2 outperforms other models by achieving a lower average mean squared error at almost all checkpoints. In the case of Syn-2, LFD-2 demonstrates top-tier performance, comparable to the two most complex models, DLinear and WNN, both in terms of parameter count and the number of input weight columns. Notably, DLinear and WNN are not feasible for training larger neural networks like CNN and DistilBERT, as discussed in the subsequent section. An interesting observation is that the error increases over the predicted steps for all models on our synthetic datasets. 
This underscores the importance of our farcasting approach and suggests strategies for determining the optimal farcasting length when designing future algorithms.

In addition we provide visual tools to qualitatively evaluate model forecasting performance. We illustrate the full trajectories for 4 farcasting models: DLinear, LFD-2, LFN, and LFS on an examplar sequence from Syn-1 with GD \footnote{See Fig. 1-3 of Appendix for more details.} and SGD update in Fig. \ref{fig:tra}. The complexity of these models increases in the order of LFD-2 < LFS < LFN < DLinear. %As shown in Figure \ref{fig:tra}, 
LFD-2 provide comparable and even better prediction to other more complex models with GD update. Trajectories with SGD updates are much more challenging for all models to handle compared to GD, which is typically used in deep learning. LFD-2 and LFN provide better predictions than DLinear and LFS in the SGD setting as they are closer to the ground truth. The trajectories of objective loss from LFD-2 on test subsets show similar converging behavior \footnote{Figure 5(b) of Appendix}. Moreover, while DLinear is effective for time series forecasting tasks, it may not perform as well for weight prediction problems, where insights from optimization are more critical than periodicity and trend. %Consequently, our more efficient approach achieves better or comparable results compared to other more complex models.

\begin{figure*}[htbp]
   % \centering
  %   \begin{subfigure}[h]{0.245\textwidth}
  %       \centering        \includegraphics[width=0.9\textwidth]{figs/DLinear_3gd2.png}
  %       \caption[Network2]%
  %       {{\small DLinear}}    
  %       % \label{fig:mean and std of net14}
  %   \end{subfigure}
  %  % \hfill
  % %  \vskip\baselineskip
  %   \begin{subfigure}[h]{0.245\textwidth}   
  %       \centering 
  %       \includegraphics[width=0.9\textwidth]{figs/LFN_3gd2.png}
  %       \caption[]%
  %       {{\small LFN}}    
  %       % \label{fig:mean and std of net34}
  %   \end{subfigure}
  %   \hfill
  %   \begin{subfigure}[h]{0.245\textwidth}   
  %       \centering 
  %       \includegraphics[width=0.9\textwidth]{figs/LFS_3gd2.png}
  %       \caption[]%
  %       {{\small LFS}}    
  %       % \label{fig:mean and std of net44}
  %   \end{subfigure}
  %   \hfill
  %   \begin{subfigure}[h]{0.245\textwidth} 
  %       \centering 
  %       \includegraphics[width=0.9\textwidth]{figs/LFD_3gd2.png}
  %       \caption[]%
  %       {{\small LFD-2}}    
  %       % \label{fig:mean and std of net24}
  %   \end{subfigure}
    % \caption[ The average and standard deviation of critical parameters ]
    % {\small Example of Farcasted Trajectories from Various Models on Syn-1 with GD Update.} 
    % \label{fig:tra_gd}
% \end{figure*}

% \begin{figure*}[h!]
    \centering
    \begin{subfigure}[h]{0.24\textwidth}
        \centering
        \includegraphics[width=0.9\textwidth]{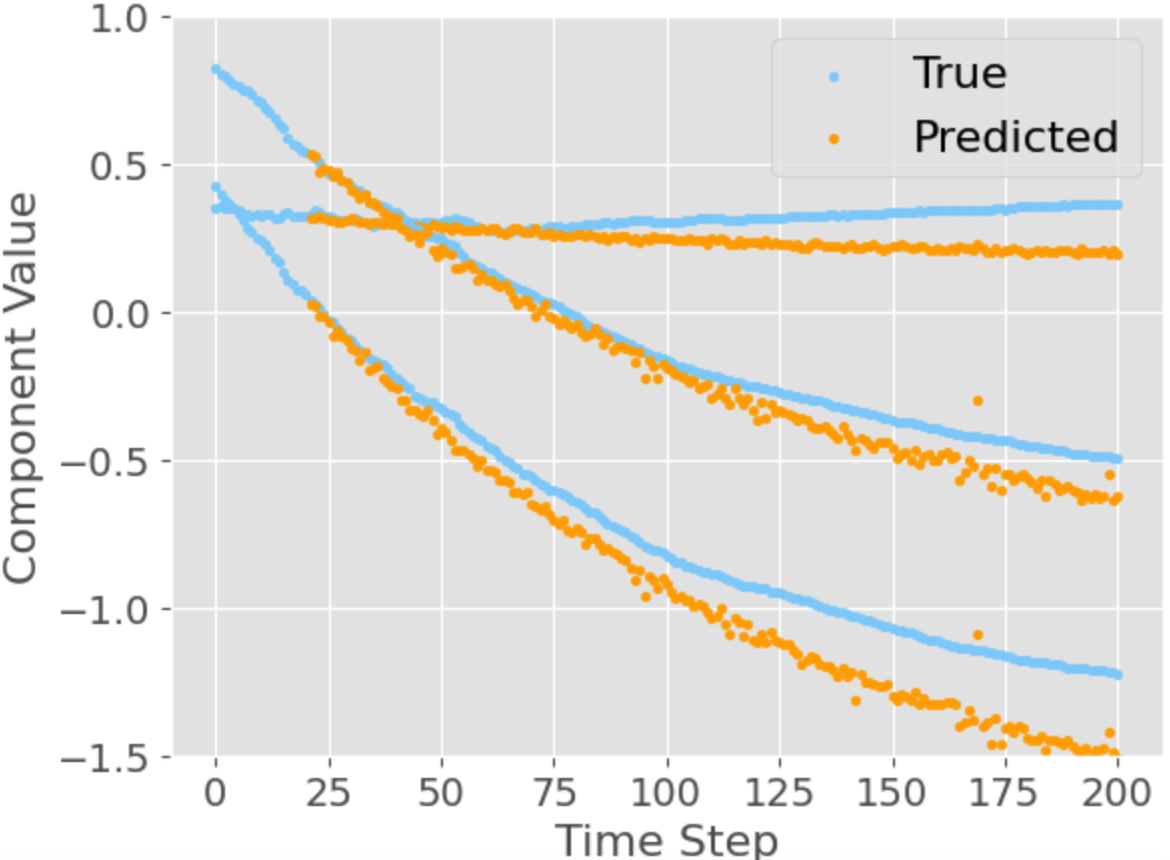}
        \caption[Network2]%
        {{\small DLinear}}    
        % \label{fig:mean and std of net14}
    \end{subfigure}
  %  \vskip\baselineskip
    \begin{subfigure}[h]{0.24\textwidth}   
        \centering 
        \includegraphics[width=0.9\textwidth]{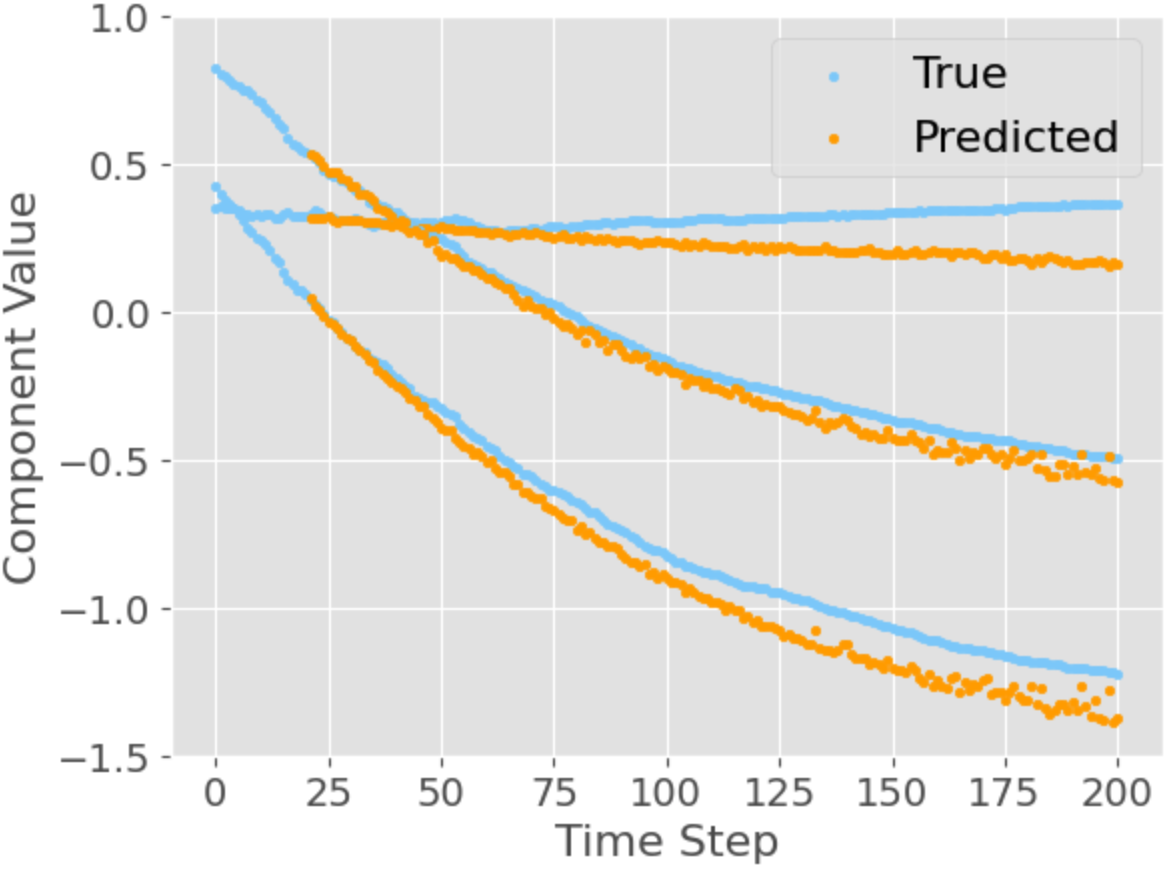}
        \caption[]%
        {{\small LFN}}    
        % \label{fig:mean and std of net34}
    \end{subfigure}
    \hfill
    \begin{subfigure}[h]{0.24\textwidth}   
        \centering 
        \includegraphics[width=0.9\textwidth]{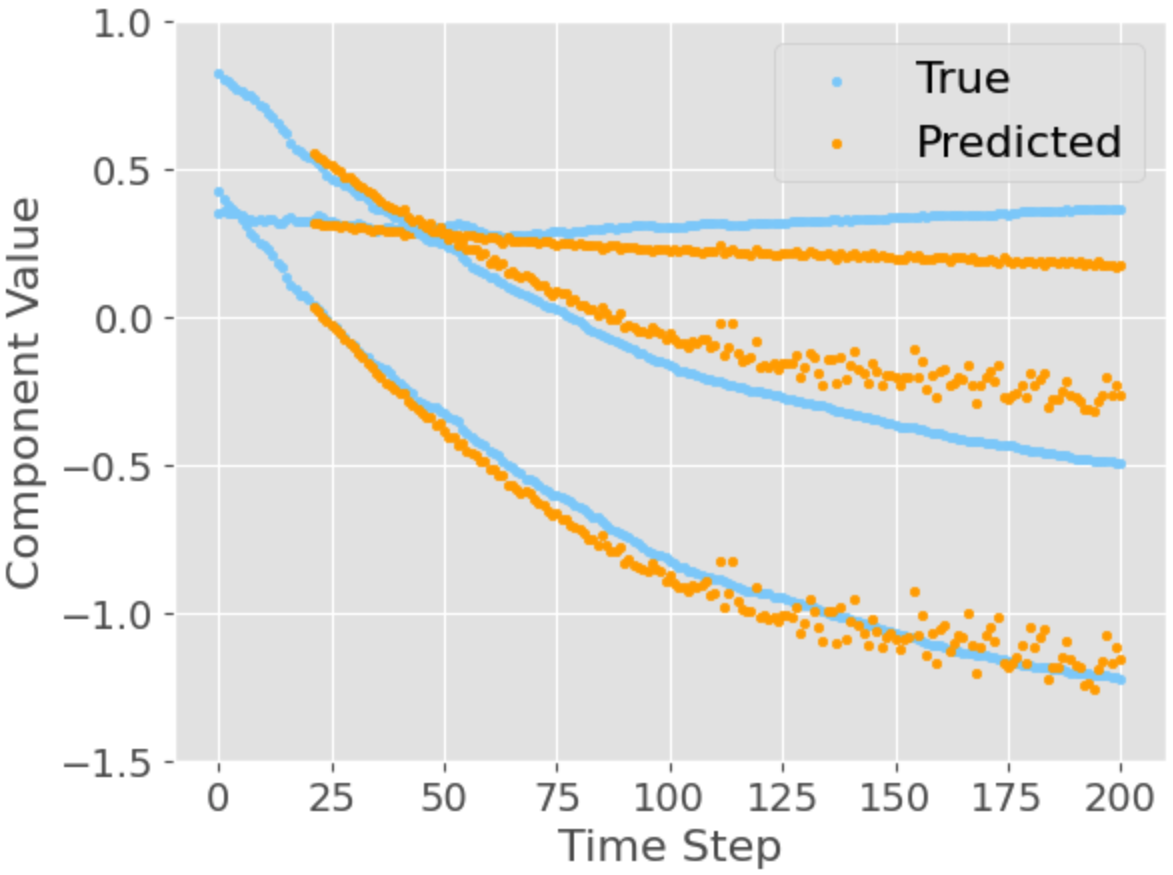}
        \caption[]%
        {{\small LFS}}    
        % \label{fig:mean and std of net44}
    \end{subfigure}
    %\hfill
    \begin{subfigure}[h]{0.24\textwidth}  
        \centering 
        \includegraphics[width=0.9\textwidth]{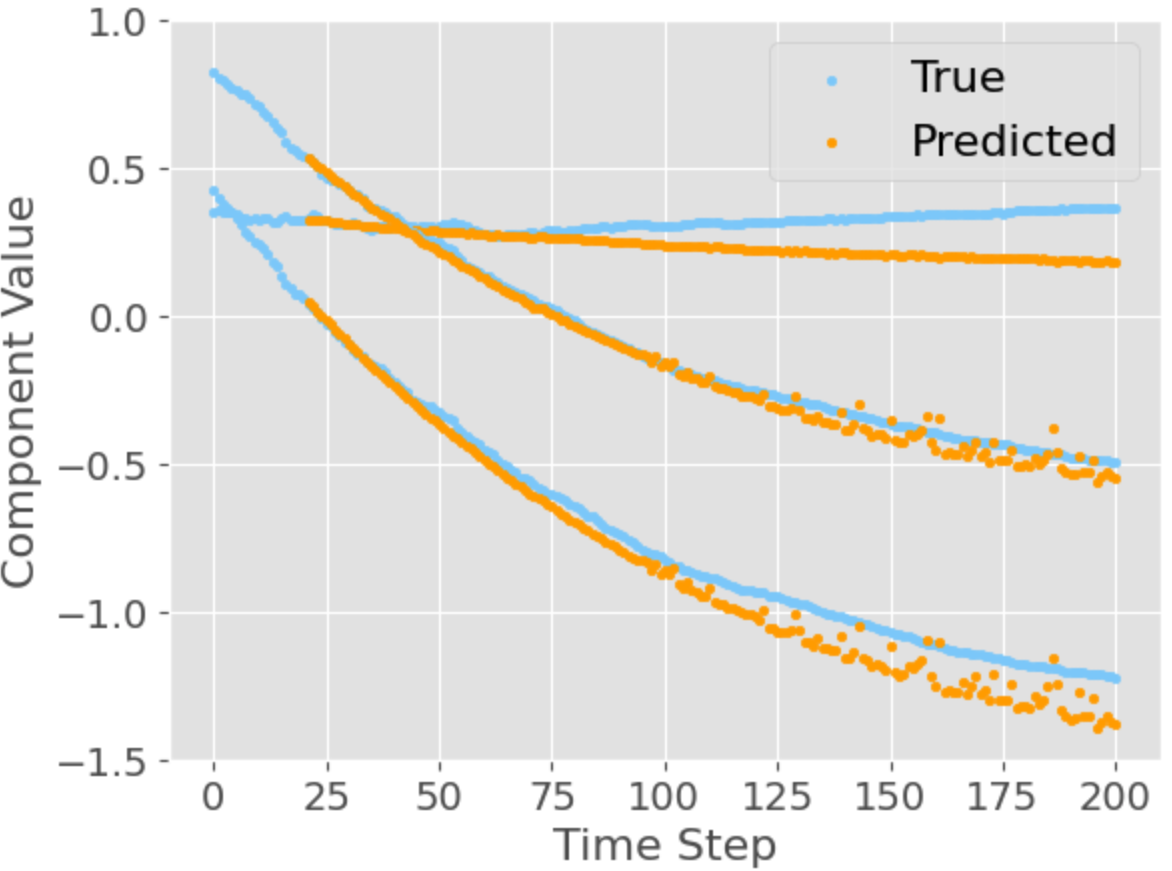}
        \caption[]%
        {{\small LFD-2}}    
        % \label{fig:mean and std of net24}
    \end{subfigure}
    \caption[ The average and standard deviation of critical parameters ]
    {\small Examples of forecasted weight sequences and their ground truth are provided for the Syn-1 synthetic data experiment, utilizing the minibatch SGD update. The feature vectors are three-dimensional, and we plot the values for each component across 200 updates for four different models (a)-(d). During the inference phase, the weight sequences from step 0 to 20 are used to forecast the sequences from 21 to 200.} 
    \label{fig:tra}
\end{figure*}

\subsection{Large Scale Deep Learning Applications}

\paragraph{Weight Sequence Generation.} We illustrate the efficiency of our model using two examples from large-scale deep learning systems. The first example is a convolutional neural network for image classification, using the standard MNIST dataset~\footnote{https://github.com/mbjoseph/pytorch-mnist/blob/master/cnn-mnist.ipynb}. The CNN model consists of two convolutional layers. We collect 10 sequences of parameter updates using the Adam optimizer while training the CNN model on the MNIST dataset with different random initializations. Each sequence contains 101 weights, and each weight is of 1,199,882 dimensions. We split the sequences into 5-2-3 for train, dev and test subsets. For each sequence, the first 11 steps of weights are used as the training sequence ($\mathbf{X}^i$), and the remaining 90 time steps are used for prediction ($\mathbf{Y}^i$).
Similarly, we perform text classification using DistilBERT on the Consumer Complaint Dataset from the Consumer Complaint Database \footnote{https://catalog.data.gov/dataset/consumer-complaint-database}. Our goal is to classify textual complaints into specific categories, such as Mortgage and Credit Reporting. We utilize an exemplar notebook~\footnote{https://github.com/vilcek/fine-tuning-BERT-for-text-classification/blob/master/02-data-classification.ipynb} that preprocesses the original dataset, uses DistilBERT embeddings, and fine-tunes a sample of the preprocessed dataset. %For more details, see the referenced notebook \footnote{https://github.com/vilcek/fine-tuning-BERT-for-text-classification/blob/master/02-data-classification.ipynb}.
The pretrained DistilBERT model produces embeddings of 66,960,393 dimensions. We generate one sequence of 60 updates using the AdamW optimizer using Amazon SageMaker  ml.g5.48xlarge instance \footnote{https://aws.amazon.com/sagemaker/pricing/}. We then split the 60 updates into 30-30 for training and testing subsets. Specifically, we train on the first 30 weight subsequence and test on the last 30 weight subsequence. The first 5 steps of weights are used as the training sequence ($\mathbf{X}^i$), and the remaining 25 time steps are used for prediction ($\mathbf{Y}^i$).

\begin{table} [h!]
% \tiny
\setlength{\tabcolsep}{4pt}
  \caption{MSE on two deep learning training applications: CNN for MNIST classification and DistilBERT for complaint classification, scaled by $10^4$. "*" indicates model outputs a single time step. Experiments were conducted on the same machine on a private server.}
  \centering
  \resizebox{1\linewidth}{!}{
  \begin{tabular}{llllllllll}
\toprule
 & \multicolumn{4}{c}{CNN with Adam} & \multicolumn{5}{c}{DistilBERT with AdamW} 
\\\cmidrule(r){2-5}\cmidrule(l){6-10} 
   Model   & 20  & 40  & 80 & 100   & 15  & 20  & 25 & 30 & sec/epoch \\
    \midrule
   Introspection & 9.59 &	\textit{.646} &	7.58	& 1.68 & .229	& .723 &	.264	& .528 &1.28*
\\
   WNN & 5.11	& 10.9	& 1.92	& 7.40 &NA	& NA& NA& NA & >10*
  \\
LFN     &.183 &	.629 &	\textit{1.37} &	\textbf{1.59}  & .109 &	\textbf{.0485} & 	.147 &	.300 & 6.75 
\\
   DLinear & NA &	NA &	NA & NA &	NA &	NA &	NA &NA & >10
    \\
   LFS &.292 &	.749	& \textit{1.37} &	1.78   & .664 &	\textit{.129} &	\textbf{.0343} &	.146 &6.89
   \\
   LFL & \textit{.179}	& .693	& 1.39	& 1.68  &\textbf{.0764}	& .134	& .377	& \textit{.0288} &3.19
   \\
  LFD-2 & \textbf{.137} &	\textbf{.599} &	\textbf{1.29}& \textit{1.60} &\textit{.0955} &	.194 &	\textit{.0803} &	\textbf{.00985} &9.42
			\\
    \bottomrule
  \end{tabular}
}
   \label{tab:dl}
\end{table} 

% \begin{table} [h!]
% % \tiny
% \setlength{\tabcolsep}{4pt}
%   \caption{MSE on two deep learning training applications: CNN for MNIST classification and DistilBERT for complaint classification, scaled by $10^4$.}
%   \centering
%   \resizebox{1\linewidth}{!}{
%   \begin{tabular}{lllllllllll}
% \toprule
%  & \multicolumn{5}{c}{CNN with Adam} & \multicolumn{5}{c}{DistilBERT with AdamW} 
% \\\cmidrule(r){2-6}\cmidrule(l){7-11} 
%    Model   & 20  & 40  & 80 & 100  &   & 15  & 20  & 25 & 30 &  \\
%     \midrule
%    Introspection & 9.59 &	\textit{.646} &	7.58	& 1.68 & 0.41*& .229	& .723 &	.264	& .528 &1.28*
% \\
%    WNN & 5.11	& 10.9	& 1.92	& 7.40&40.63* &NA	& NA& NA& NA & NA
%   \\
% LFN     &.183 &	.629 &	\textit{1.37} &	\textbf{1.59} &20.61 & .109 &	\textbf{.0485} & 	.147 &	.300 & 6.75 
% \\
%    DLinear & NA &	NA &	NA &	NA & NA &	NA &	NA &	NA &NA & NA
%     \\
%    LFS &.292 &	.749	& \textit{1.37} &	1.78 & 13.6  & .664 &	\textit{.129} &	\textbf{.0343} &	.146 &6.89
%    \\
%    LFL & \textit{.179}	& .693	& 1.39	& 1.68 & 8.11 &\textbf{.0764}	& .134	& .377	& \textit{.0288} &3.19
%    \\
%   LFD-2 & \textbf{.137} &	\textbf{.599} &	\textbf{1.29}& \textit{1.60} & 22.65&\textit{.0955} &	.194 &	\textit{.0803} &	\textbf{.00985} &9.42
% 			\\
%     \bottomrule
%   \end{tabular}
% }
%    \label{tab:dl}
% \end{table} 

\paragraph{Results.} We present the mean squared error (MSE) results for seven models at various checkpoints in Table \ref{tab:dl}. The data size demands a high level of memory and computing resources, particularly for large language models like DistilBERT. For instance, storing a sequence of 60 weights requires approximately 40GB of memory. These challenges cause models such as WNN and DLinear with complex transformation operations for training to fail. On the other hand, our proposed model significantly outperforms others in predicting future weights, achieving notably lower MSE. For example, forecasting the $30^{th}$ step using DistilBERT results in an order of magnitude improvement over most models while remaining the same order of magnitude of computing time. 

\subsection{Ablational Experiments}
We address the following two questions via ablational experiments in this section.
\begin{table}[htbp]
\tiny
% \begin{minipage}[t]{1\linewidth}
\setlength{\tabcolsep}{2.5pt}
  \caption{Farcast with loss, scaled by $10^4$.}
  \centering
  \resizebox{0.9\linewidth}{!}{
  \begin{tabular}{lllll}
\toprule
   Experiment  &  40 & 80 & 160 & 200   \\
    \midrule
   Syn-1 GD & .196(.015)	& 4.42 (.47) & 43.8 (5.5) &	78.0(6.7) \\
   Syn-1 SGD & 10.9(1.3)	& 52.9(8.5)& 194(23)	& 278(18)\\
   Syn-2 SGD &  1.74(.12)	&24.7(1.5)&	126(9)&	175(5)  \\
   Syn-2 Adam & .799(.229)	&13.5(5.7)	&37.4(4.4)	&54.7(11.1)	\\
    \bottomrule
  \end{tabular}
  }
   \label{tab:loss}
\end{table} 
% We address the following two questions via ablational experiments in this section.
% \paragraph{Can loss help farcasting?} In ML/DL loss at each update is accessible. For example, in DL this is through a forward pass. We are interested in answering the question if we augment loss to the weights i.e. ${(\mathbf{w}_i,l_i)}$ to help our predict future weights. So we augument such losses in our syn-1 and syn-2 experiments. Results in \ref{tab:loss} show that our model LFD-2 does not benefit significantly from adding such losses. A possible explanation a linear mapping may not be adequate for learning the loss function with repective to weight; furthermore, the loss may not directly benefit the prediction of weight but it may directly benefit the prediction of future loss. More complex architecture or weight topology needed to guide the learning of weights using loss. 
\paragraph{Can training loss help the forecast?} In ML and DL, the loss at each update is readily accessible, typically through a forward pass in deep learning. We are interested in exploring whether augmenting the weights with the loss values, i.e., using ${(\mathbf{w}_i, l_i)}$, can improve the prediction of future weights. Therefore, we incorporated such losses in our Syn-1 and Syn-2 experiments. The results in Table \ref{tab:loss} indicate that our model, LFD-2, does not benefit significantly from adding these losses. A possible explanation is that a linear mapping may not be sufficient for effectively learning the relationship between the loss function and the weights. Moreover, while the loss may not directly enhance the prediction of weights, it could directly benefit the prediction of future losses. More complex architectures or weight topologies might be necessary to leverage loss information for guiding the learning of weights.

% \end{minipage}
   % \vskip\baselineskip
% \begin{minipage}[t]{1\linewidth}
\begin{table}[htbp]
\tiny
\setlength{\tabcolsep}{2.5pt}
  \caption{Farcast with $\{\mathbf{w}_0,\mathbf{w}_1\}$, scaled by $10^4$.}
  \centering
  \resizebox{0.9\linewidth}{!}{
  \begin{tabular}{lllll}
\toprule
   Experiment  &  40 & 80 & 160 & 200   \\
    \midrule
   Syn-1 GD &247(15)&1625(85)	&4765(220)	&6017(190)\\
   Syn-1 SGD &230(14)	&1486(103) &4419(188)	&5710(279)\\
   Syn-2 SGD & 12.6(1.1)
   &75.2(4.4)	&235(24)	&294(32) \\
   Syn-2 Adam & 14.5(2.6)&	72.5(15.2)	&163(27)	&186(29)	\\
    \bottomrule
  \end{tabular}
  }
   \label{tab:grad}
% \end{minipage}
\end{table} 

% \paragraph{How many gradient steps should be performed to get a good predictive performance?} When the number of parameters are large such as in LLM (i.e. DistilBERT), we want to limit the number of gradient steps (or backward pass) to predict. We perform such an experiment on Syn-1 and Syn-2 to serve as a proxy to LLM such as DistilBERT. We perform experiment on Syn-1 and Syn-2 datasets where we only use $\{\mathbf{w}_0,\mathbf{w}_1\}$ to forecast $\{\mathbf{w}_i\}$ for $i \in \{2,3,...,200\}$. The results from our model LFD-2 shown in Table \ref{tab:grad} indicate that linear regression with least square fit without neural network, works much worse than linear regression with neural network. An possible explanation is that weights in neural network do not change significantly over time \cite{sinha2017introspection}. It is not too bad to use the first 2 steps and forecast not too many steps ahead; however the performance does decay in this case compared with the case we choose to train with first 21 steps. In practice. an analyst may decide to train some number of steps and then using our model LFD-2 to forecast. 

\paragraph{How many gradient steps are necessary for good predictive performance?} When dealing with models that have a large number of parameters, such as large language models (LLMs) like DistilBERT, it is important to limit the number of gradient steps (or backward passes) needed to reduce training cost. To investigate this, we conducted experiments using the Syn-1 and Syn-2 datasets as proxies for LLMs such as DistilBERT. Specifically, we used $\{\mathbf{w}_0, \mathbf{w}_1\}$ to forecast ${\mathbf{w}_i}$ for $i \in \{2, 3, \ldots, 200\}$. The results from our model LFD-2, shown in Table \ref{tab:grad}, indicate that linear regression with a least squares fit performs significantly worse than that trained by a neural network. One possible explanation, as suggested by previous work \cite{sinha2017introspection}, is that the weights in neural networks generally do not change significantly over time. While using only the first two steps and forecasting a limited number of steps ahead is not overly detrimental, performance does decline compared to training with the first 21 steps in Table \ref{tab:syn1} and \ref{tab:syn2}. In practice, an analyst may choose to train for a certain number of steps and then use our model LFD-2 for forecasting, depending on computational resources.

\section{Conclusion}
We address the computational challenges of training ML and DL systems with a novel framework. Our proposed approach adapts long-term time series forecasting techniques to neural network weight predicting problem and we further improve efficiency by using minimal number of previous steps while maintaining superior accuracy. Adaptive weight prediction or simplification are potential directions for future work. While we validate our method on CNNs and DistilBERT, scaling to larger architectures like full BERT, or even GPT models remains an exciting direction. %Given that farcasting is lightweight, it could be particularly useful in LLM pretraining, where computational savings are most valuable. However, additional experiments on these architectures are necessary to fully validate scalability.

\end{document}